\documentclass[pdflatex,sn-mathphys-num]{sn-jnl}

\usepackage{array}%
\usepackage{graphicx}%
\usepackage{makecell}%
\usepackage{lineno}%
\usepackage{amsfonts}%
\usepackage{multirow}%
\usepackage{amsmath,amssymb,amsfonts}%
\usepackage{amsthm}%
\usepackage{mathrsfs}%
\usepackage[title]{appendix}%
\usepackage{xcolor}%
\usepackage{textcomp}%
\usepackage{manyfoot}%
\usepackage{booktabs}%
\usepackage{algorithm}%
\usepackage{algorithmicx}%
\usepackage{algpseudocode}%
\usepackage{listings}%

\theoremstyle{thmstyleone}%
\theoremstyle{thmstyletwo}%

\theoremstyle{thmstylethree}%

\usepackage{geometry}
\begin{document}

\title[Article Title]{Let Robots Feel Your Touch: Visuo-Tactile Cortical Alignment for Embodied Mirror Resonance}


\author[1,4]{\fnm{Tianfang} \sur{Zhu}}\email{fun0515@foxmail.com}

\author[1]{\fnm{Ning} \sur{An}}\email{anningnm@icloud.com}

\author[1]{\fnm{Rui} \sur{Wang}}\email{wangrui@air.tsinghua.edu.cn}
\author[3]{\fnm{Jiasi} \sur{Gao}}\email{gaojiasi@buaa.edu.cn}

\author*[2]{\fnm{Qingming} \sur{Luo}}\email{qluo@hainanu.edu.cn}

\author*[2,4]{\fnm{Anan} \sur{Li}}\email{aali@hainan.edu.cn}
\author*[1]{\fnm{Guyue} \sur{Zhou}}\email{zhouguyue@air.tsinghua.edu.cn}

\affil[1]{\orgdiv{Institute for AI Industry Research}, \orgname{Tsinghua University}, \orgaddress{\city{Beijing}, \country{China}}}

\affil[2]{\orgdiv{Key Laboratory of Biomedical Engineering of Hainan Province, School of Biomedical Engineering}, \orgname{Hainan University}, \orgaddress{\city{Haikou}, \country{China}}}

\affil[3]{\orgdiv{School of New Media Art and Design}, \orgname{Beihang University}, \orgaddress{\city{Beijing}, \country{China}}}

\affil[4]{\orgdiv{MoE Key Laboratory for Biomedical Photonics, Wuhan National Laboratory for Optoelectronics}, \orgname{Huazhong University of Science and Technology}, \orgaddress{\city{Wuhan}, \country{China}}}


\abstract{Observing touch on another's body can elicit corresponding tactile sensations in the observer, a phenomenon termed mirror touch that supports empathy and social perception.
This visuo-tactile resonance is thought to rely on structural correspondence between visual and somatosensory cortices, yet robotic systems lack computational frameworks that instantiate this principle.
Here we demonstrate that cortical correspondence can be operationalized to endow robots with mirror touch.
We introduce Mirror Touch Net, which imposes semantic, distributional and geometric alignment between visual and tactile representations through multi-level constraints, enabling prediction of millimetre-scale tactile signals across 1,140 taxels on a robotic hand from RGB images.
Manifold analysis reveals that these constraints reshape visual representations into geometry consistent with the tactile manifold, reducing the complexity of cross-modal mapping.
Extending this alignment framework to cross-domain observations of human hands enables tactile prediction and reflexive responses to observed human touch.
Our results link a neural principle of visuo-tactile resonance to robotic perception, providing an explainable route towards anticipatory touch and empathic human-robot interaction. 
Code is available at \url{https://github.com/fun0515/Mirror-Touch-Net}.
}




\maketitle

\section{Introduction}\label{sec1}
\label{sec:intro}
\begin{figure}[tbp]
    \centering
    \includegraphics[width=0.999\textwidth]{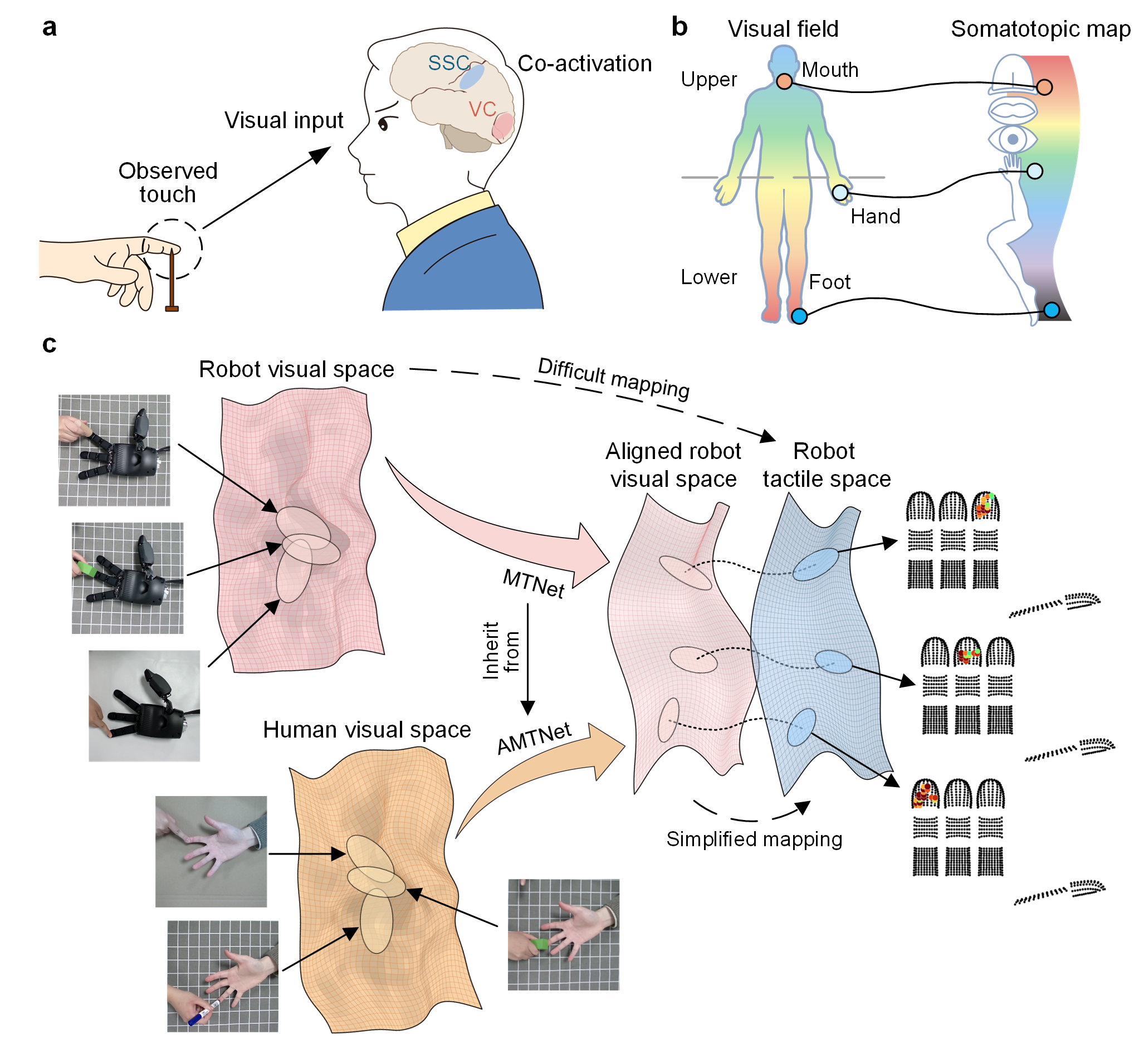}
    \caption{\textbf{From cortical alignment to robotic mirror touch.}
    \textbf{a,} Observed touch on another body may co-activate the observer’s visual cortex (VC) and somatosensory cortex (SSC), eliciting a similar sensory response.
    \textbf{b,} This visuo-tactile resonance may be supported by structural alignment between visual and somatosensory maps (adapted from~\cite{VC_alignment_2026N}).
    \textbf{c,} Inspired by cortical alignment, the proposed robotic mirror touch framework recasts tactile inference as visuo–tactile representation alignment.
    MTNet establishes a compact robotic alignment to simplify cross-modal mapping, while AMTNet transfers this alignment to images of human touch, enabling robotic tactile resonance with observed human touch.
    Robotic setup is detailed in Fig.~\ref{fig:supply_setup}.
    }
    \label{fig:intro}
\end{figure}

Mirror touch provides a striking example of how perception can be transformed into a bodily representation, linking social perception to embodied cognition. As Fig.~\ref{fig:intro}a depicts, observing another individual being touched can recruit a corresponding tactile state on the observer’s own body~\cite{observation_touch_2004Neuron, Somatosensation_2010NN, feel_2009Plos}, a phenomenon often referred to as mirror touch~\cite{mirror_touch_2005Brain,mirror_touch_2015CN} and, more broadly, visuo-tactile resonance.
This capacity is important for actions requiring the inference of expected tactile consequences~\cite{mirror_touch_action_2001NRN}, and it further supports key aspects of social cognition, including affective touch~\cite{affective_touch_2023JN}, pain observation~\cite{pain_empathy_2004Science}, and empathy in caregiving and cooperative contexts~\cite{mirror_touch_2007NN,mirror_touch_2018CN}. 
The widespread presence of visually induced tactile processing in humans indicates that it is a basic organizational principle governing the integration of visual and somatosensory information in the brain, rather than an incidental feature of sensory processing.
Therefore, characterizing this neural architecture is not only a fundamental question in neuroscience but also a prerequisite for its formalization and application in building artificial systems capable of visuo-tactile resonance.

At the cortical level, visuo-tactile resonance is supported by a structural alignment of visual and somatosensory representations. 
Neuroimaging studies show that observing touch engages regions that overlap with those activated by actual touch~\cite{observation_touch_2004Neuron,Somatosensation_2010NN}, primarily the secondary somatosensory cortex and occasionally the primary somatosensory cortex. 
Visual and tactile signals converge in the parietal cortex, where they are integrated into shared topographic maps linking external visual space with internal bodily coordinates~\cite{PC_integration_2006NN,PC_integration_2010EJN,PC_integration_2023NC}. 
A recent cross-modal mapping study provides direct evidence for such alignment between visual and somatosensory body representations~\cite{VC_alignment_2026N}, showing that body-position tuning is systematically related to visual-field tuning and predicts preferences for both visual-field locations and body-part categories.
For example, tuning to the upper visual field coincides with somatotopic regions encoding higher-elevation body parts (Fig.~\ref{fig:intro}b).
This cortical alignment suggests that visuo-tactile resonance is grounded in a strongly constrained mapping, rather than a loose statistical association, and may therefore reflect a structured computational principle.

Robotic systems, however, currently lack mechanisms that instantiate this alignment principle. 
The growing recognition of tactile sensing for dexterous manipulation has motivated diverse tactile sensing modalities and their integration into embodied robots~\cite{tactile_sensor_2020RAL,tactile_sensor_2021SR,tactile_sensor_2024ICRA,tactile_sensor_2026NS}, particularly anthropomorphic robotic hands~\cite{tac_hand_2025SciAdv,tac_hand_2005NMI,tac_hand_2025NC}, for fine-grained contact estimation.
Yet the dominant paradigm still treats touch primarily as a passive feedback signal: vision guides action selection, while tactile feedback refines motor commands~\cite{TacMani_2019SciRob,TacMani_2023CoRL,TacMani_2025CoRL,TacVisionMani_2025arxiv,TacMani_2026RAL}. 
Tactile signals are available only when physical interaction occurs, introducing temporal delays that prevent robots from performing the anticipatory tactile inference that humans routinely achieve by observing objects before contact. 
Although recent studies have adopted variational autoencoders~\cite{OmniVTA} and masked autoencoders~\cite{tac_pred_MAE_2025CoRL,tac_pred_MAE_2025ICLR,tac_pred_MAE_2026SciRob} to predict tactile signals from visual inputs, these methods are trained mainly by minimizing reconstruction losses, leaving them to learn a difficult cross-modal mapping without the structured alignment between visual and tactile representations observed in the human brain. 
This contrast naturally raises a fundamental question: \textbf{Can cortical alignment principles be translated into a computational framework to endow embodied systems with mirror touch?}

Here we address this question affirmatively by proposing Mirror Touch Net (MTNet), an architecture that translates cortical alignment principles into multi-level computable constraints for enabling visuo-tactile resonance in robotic systems.
Following the brain’s sensory segregation and integration, MTNet models visual and tactile modalities in separate representational spaces, with their structured alignment enforced through six constraints, including distributional alignment, semantic correspondence, and sample-wise distance consistency.
From RGB observations alone, it predicts dense tactile signals over 1,140 independent taxels on a robotic hand, achieving spatial resolution comparable to human skin while spanning a larger cross-modal gap than prior vision-based or sparse tactile approaches.
Manifold analysis shows that these constraints regularize otherwise disordered visual representations into a structured geometry consistent with the tactile manifold. Without such enforced alignment, the model is left to learn a difficult nonlinear mapping across modalities, resulting in sharp performance degradation.
Crucially, we generalize this principle beyond the robot's own morphology by extending alignment from cross-modal to cross-domain correspondence. 
This leads to Adaptive MTNet (AMTNet), which transfers tactile inference across morphologies to observed human touch solely through alignment in the visual space. 
Physical experiments validate this capability: camera-observed contact on a human hand elicits corresponding tactile activations and reflexive motor responses in the robot, reproducing the core phenomenology of mirror touch.
These results demonstrate that embodied machines can perceive and respond to others’ tactile experiences through cortical alignment principles, with implications for human-robot interaction and neurally grounded embodied intelligence.

\section{Results}\label{sec2}
\subsection{Experimental framework for visuo-tactile resonance}\label{sec:result_intro}

To investigate whether observed touch can be transformed into a corresponding tactile state in a robotic system, we built an embodied platform for a mirror-touch-like setting (Fig.~\ref{fig:supply_setup}a).
In this setup, a contact event is captured by a monocular RGB camera and used to infer tactile states on a robotic hand without contact. 
Because no depth sensing or motion capture is used, tactile consequences have to be inferred solely from RGB image sequences. 
Tactile sensing is provided by a four-fingered dexterous hand equipped with distributed electromagnetic sensors (Fig.~\ref{fig:supply_setup}b-c). 
This configuration differs from common vision-based tactile sensors~\cite{vt_cl_2025ICRA,OmniVTA}, in which visuo-tactile mapping is often formulated as image-to-image translation.
By contrast, our hand contains 1,140 discrete mechanoreceptor-like taxels across 11 finger segments.
The model is therefore required to predict, from consecutive image frames, the three-dimensional (3D) force vector for each taxel at the current time step.
With a 1-millimetre spatial resolution is close to that of human skin~\cite{human_tactile_resolution_1981,human_tactile_resolution_1994}, the tactile array creates, together with the dense RGB visual space, a substantial modality gap between visual and tactile sensing.

Using this experimental framework, we formulated visuo-tactile resonance as two progressive computational challenges (Fig.~\ref{fig:intro}c).
The first is a cross-modal prediction task in which the system must infer tactile contact signals over the robotic hand surface in step with the visual input from recorded image sequences of contact events. 
The second requires extending the robotic mapping to observations of human hands through cross-domain transfer, enabling the generation of corresponding tactile responses across morphological differences. 
Together, these challenges arise from two sources: a cross-modal gap between dense visual observations and sparse tactile signals, which introduces many-to-one visual-to-tactile ambiguities, and a cross-domain supervision gap due to the absence of tactile ground truth for human touch. 
We therefore address both challenges through the alignment principle, implemented by the network architecture and learning constraints described below.

\begin{figure}[tbp]
    \centering
    \includegraphics[width=0.99\textwidth]{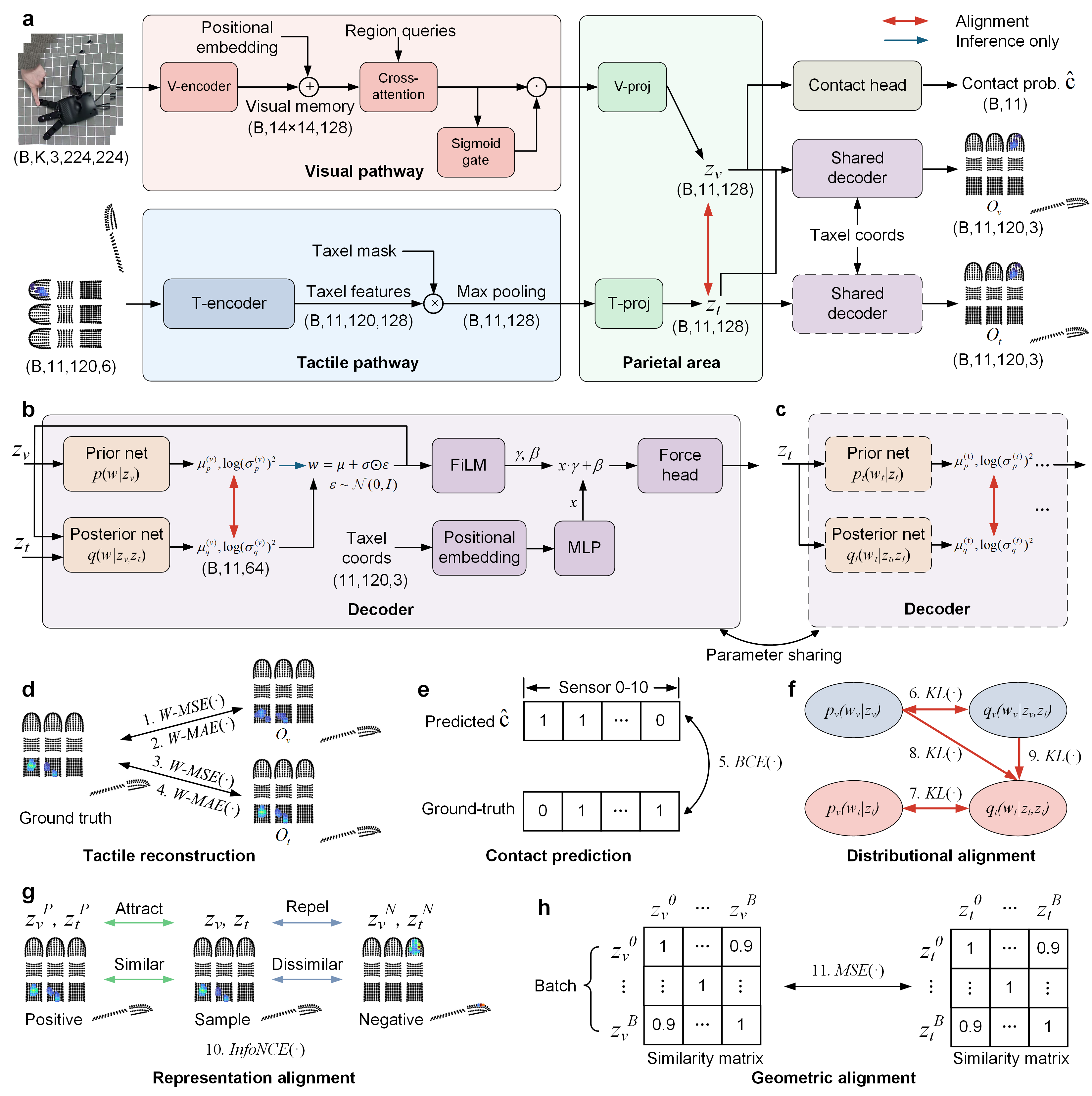}
    \caption{\textbf{Architecture and training objectives of MTNet.}
    \textbf{a,} Visual and tactile inputs are processed by separate pathways and then projected to a parietal-like area. A shared decoder maps the resulting latent features to tactile outputs.
    \textbf{b,} Visual features drive tactile prediction, while paired tactile features condition the decoder posterior during training; inference uses the visual prior distribution alone.
    \textbf{c,} Tactile features are decoded directly to anchor the shared decoder.
    \textbf{d-h,} Eleven training objectives for optimizing MTNet.
    \textbf{d,} Four taxel-level reconstruction losses. W-MSE and W-MAE denote contact region weighted mean squared error (MSE) and mean absolute error (MAE).
    \textbf{e,} Sensor-level contact state classification loss computed using binary cross-entropy (BCE).
    \textbf{f,} Four KL losses, including two prior posterior regularizers within the visual and tactile pathways and two cross-modal terms anchored by the tactile posterior.
    \textbf{g,} InfoNCE~\cite{infonce} style semantic loss that pulls visual and tactile features of similar contact events closer.
    \textbf{h,} Geometric alignment loss that preserves sample distance structure across modalities. Details are provided in Methods.
    }
    \label{fig:mtnet}
\end{figure}

\subsection{Cortically inspired alignment architecture}\label{sec:result_mtnet}
To model human’s visuo-tactile inference, we propose MTNet, a cortically inspired architecture that aligns visual observations with tactile contact (Fig.~\ref{fig:mtnet}a).
Following cortical multisensory organization, MTNet adopts a dual-stream architecture: the visual pathway extracts features from image sequences, whereas the parallel tactile pathway processes contact signals formulated as point clouds. 
Notably, both encoders employ lightweight backbones (e.g., ResNet and PointNet) to ensure that MTNet's efficacy is driven by its architectural inductive biases rather than massive model capacity. 
Analogous to multisensory convergence in parietal cortices, these two distinct streams are projected into a unified latent space before being mapped back to original tactile domain via a shared decoder (Fig.~\ref{fig:mtnet}b-c).
The tactile encoding and generative decoding processes are conditioned on the 3D coordinates of taxels, which impose an artificial somatotopic map and ground the latent representation in robotic hand morphology.

Learning such a cross-modal mapping from scratch is inherently challenging due to the semantic and geometric gap between visual observations and tactile contact.
Therefore, MTNet is trained with a composite objective consisting of 11 carefully designed loss terms, including five reconstruction terms and six cross-modal alignment constraints.
The five reconstruction losses ensure basic predictive accuracy at both the taxel and sensor levels (Fig.~\ref{fig:mtnet}d-e). 
Beyond minimizing numerical errors over 1,140 taxels, MTNet jointly predicts the binary contact states of 11 tactile sensors and uses them to spatially modulate taxel-level predictions. 
By assigning higher loss weights to activated regions, a contact-aware mechanism forces the model to prioritize the optimization of actual contact events rather than persistent non-contact background regions.

\begin{figure}[tbp]
    \centering
    \includegraphics[width=0.99\textwidth]{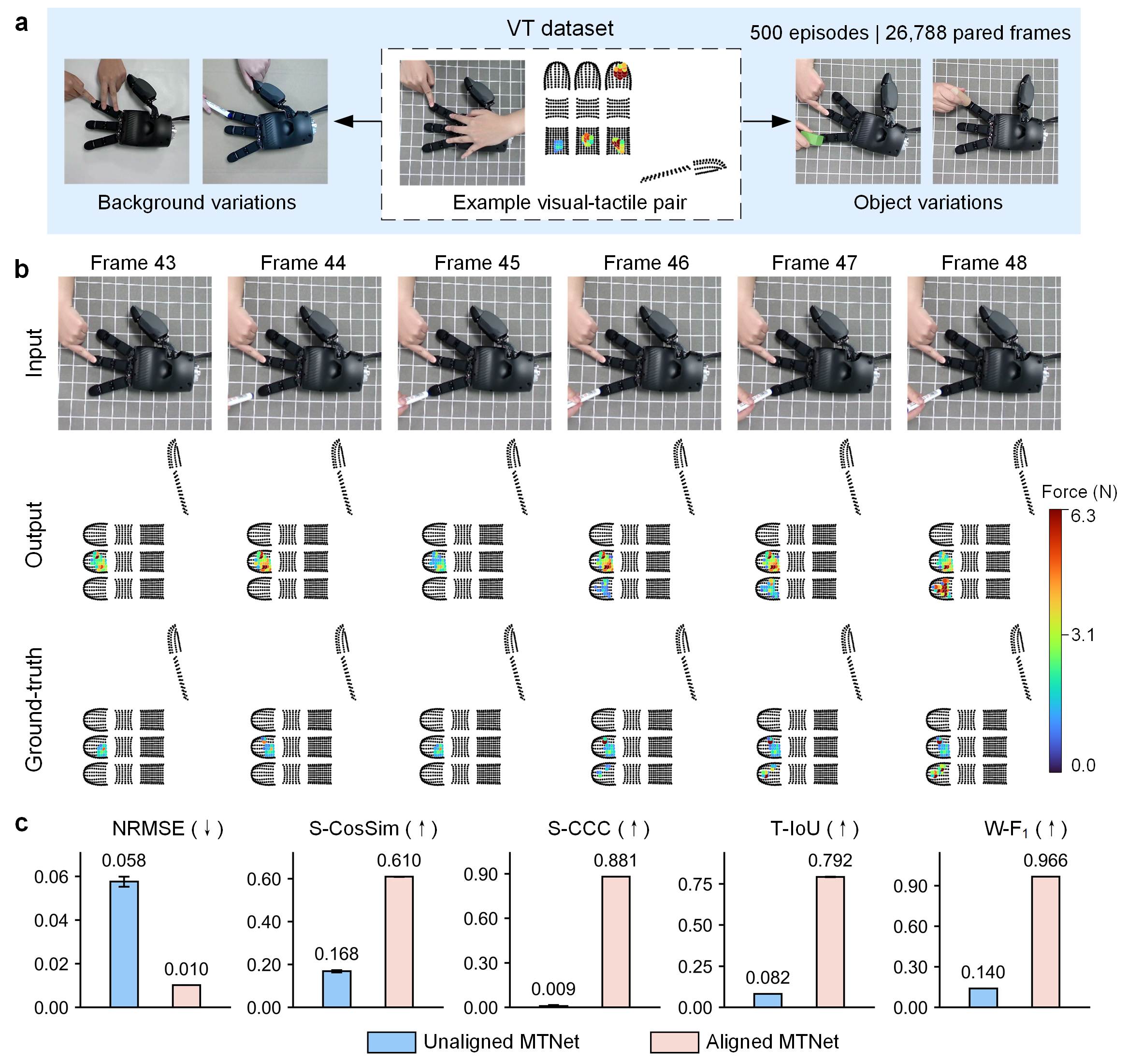}
    \caption{\textbf{Tactile prediction performance of MTNet.}
    \textbf{a, }Paired robotic hand visual-tactile dataset used for MTNet training and evaluation.
    \textbf{b, }Representative MTNet tactile prediction from $K=5$ consecutive visual frames; only the final, temporally aligned visual input is shown for clarity.
    Scatter colors denote the resultant force magnitude at each taxel.
    \textbf{c, }Quantitative comparison of MTNet and its variant without alignment constraints across five tactile prediction metrics (mean $\pm$ SD over five runs). $\uparrow$, higher is better; $\downarrow$, lower is better. Metrics are detailed in Methods.
    }
    \label{fig:mtnet_result}
\end{figure}

However, precise reconstruction is largely unattainable without bridging the underlying modality gap.
The core innovation of MTNet lies in six alignment constraints, which enforce structural consistency between the visual and tactile modalities at the probabilistic, feature, and geometric levels, thereby discouraging the model from learning brittle numerical shortcuts.
At the probabilistic level (Fig.~\ref{fig:mtnet}f), MTNet adopts a Conditional Variational Autoencoder (CVAE)~\cite{cvae} formulation and regularizes the latent space with four complementary Kullback-Leibler (KL) divergence terms, including both intra-modal prior-posterior regularization and cross-modal KL losses that anchor the visual prior and posterior to a stabilized moving-average estimate of the tactile posterior.
At the feature level (Fig.~\ref{fig:mtnet}g), a force-aware contrastive objective aligns cross-modal representations by explicitly pulling representations of similar physical interactions together while pushing disparate events apart.
At the geometric level (Fig.~\ref{fig:mtnet}h), a relational constraint aligns batch-wise geometry by penalizing discrepancies between the pairwise distance matrices of visual and tactile samples, encouraging physically similar touch events to occupy similar positions in the visual latent space.
Together, these constraints reshape the visual manifold to align with tactile structure, providing a structured latent basis for more reliable visual-to-tactile prediction.

\subsection{Visual-to-tactile prediction through alignment}\label{sec:mtnet_result}
To systematically evaluate the predictive capabilities generated by alignment constraints, we construct a dedicated robotic visual-tactile dataset (VTDataset) comprising 500 interactive episodes, yielding 26,788 paired visual and tactile frames (Fig.~\ref{fig:mtnet_result}a).
When evaluated on this dataset, MTNet demonstrates effective cross-modal prediction.
Although visual occlusion makes it unrealistic to exactly predict the force value at every individual taxel, the model successfully inferred the sensor-level spatial distribution of contacts and estimate approximate force magnitudes across different objects (Fig.~\ref{fig:mtnet_result}b).

To quantify prediction performance, we develop an evaluation framework with five metrics: Normalized Root Mean Square Error (NRMSE), Spatial Cosine Similarity (S-SimCos), Smooth Concordance Correlation Coefficient (S-CCC), Temporal Intersection over Union (T-IoU), and Windowed F1 score (W-F1). 
These metrics characterize magnitude errors, force distributions, force evolution, contact timing, and are tailored to sparse and localized tactile contacts to prevent irrelevant non-contact signals from skewing the evaluation (see Methods and Fig.~\ref{fig:supply_metrics}). 
Specifically, NRMSE and S-SimCos measure the absolute numerical errors and the spatial distribution congruence of the predicted force vectors; S-CCC quantifies the temporal consistency of force variations during continuous interactions; T-IoU and W-F1 further assess the accuracy of binary contact event detection, respectively.

Based on these tailored metrics, we conduct a quantitative ablation study comparing the full MTNet against a baseline model trained only with five reconstruction losses (i.e., without six alignment constraints). 
As shown in Fig.~\ref{fig:mtnet_result}c, the full MTNet consistently outperforms the unconstrained baseline across all metrics.
The supplementary ablation results in Tab.~\ref{tab:alignment_ablation} further show that removing any individual alignment constraint degrades performance compared with the fully aligned MTNet.
Visualizations of the unconstrained variant reveal spatially inconsistent and physically implausible tactile predictions, whereas the full MTNet accurately localizes contact timing and regions (Fig.~\ref{fig:supply_MTNetResult}).
Notably, the training losses of both model variants decrease normally (Fig.~\ref{fig:supply_loss_curve}), indicating that the poor performance of the unconstrained baseline does not result from optimization collapse. 
Rather, since many temporal and spatial locations may correspond to non-contact states, reconstruction losses can be reduced by fitting zero values and mean responses. 
The alignment constraints therefore provide essential cross-modal supervision, anchoring latent representations to contact-relevant structures and enabling accurate sensor-level predictions.
These results confirm that cross-modal prediction based solely on numerical regression is difficult and that the alignment principle directly improves tactile prediction, as further supported by manifold analysis in the following section.

\begin{figure}[tbp]
    \centering
    \includegraphics[width=0.99\textwidth]{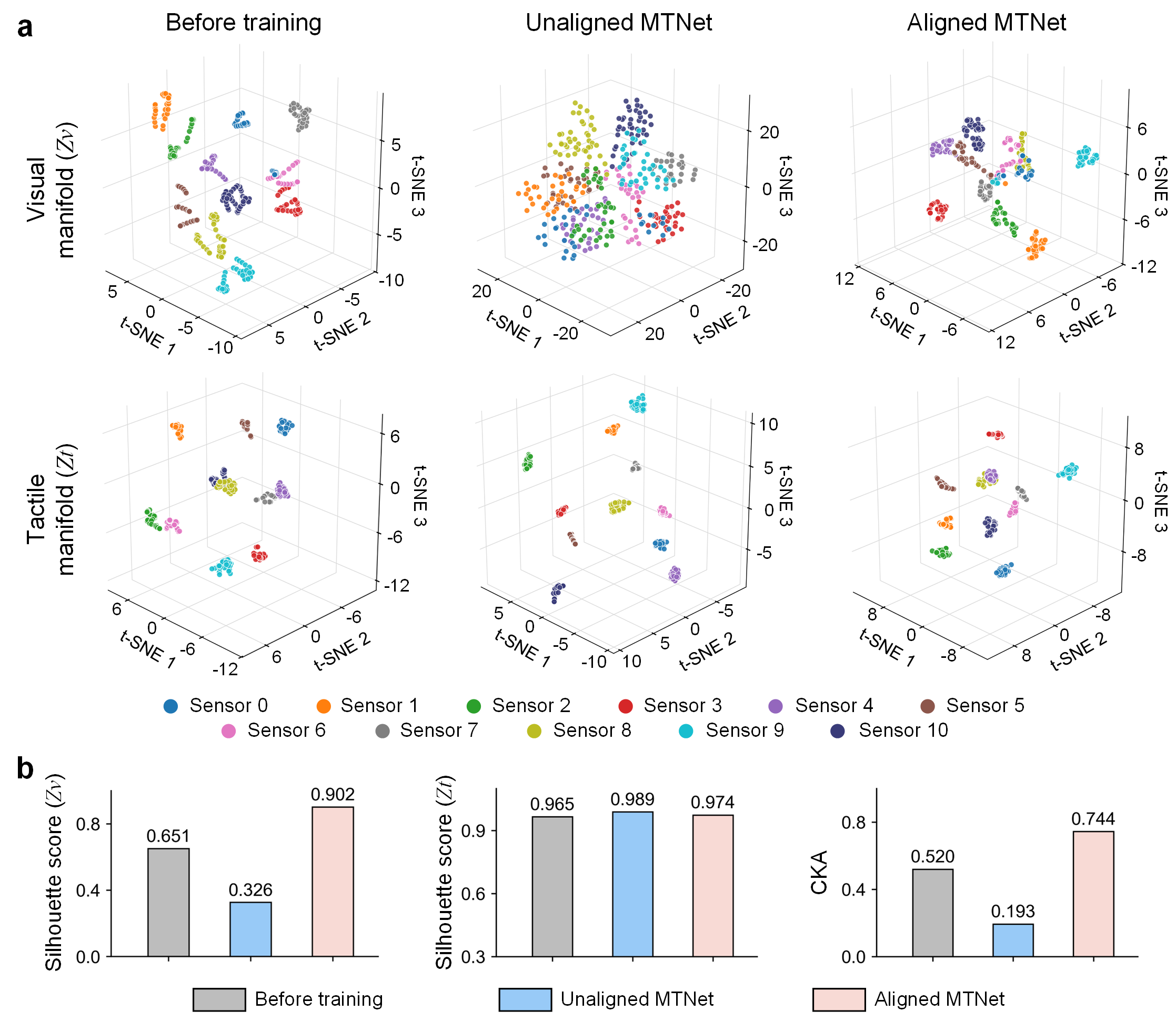}
    \caption{\textbf{Manifold analysis of alignment constraints.}
    \textbf{a, }t-SNE visualizations of visual features and tactile features ($Z_v$ and $Z_t$ in Fig.~\ref{fig:mtnet}a) extracted from contact frames during isolated touch events across 11 tactile sensors for three model variants: untrained MTNet, MTNet without alignment constraints, and fully trained MTNet.
    \textbf{b, }Quantitative comparison of silhouette coefficient and centered kernel alignment (CKA) across the three models.
    }
    \label{fig:mtnet_manifold}
\end{figure}

\subsection{Manifold analysis of cross-modal alignment}
\label{sec:manifold}
To assess the role of alignment constraints in cross-modal prediction, we perform manifold analysis of visual and tactile representations.
These latent features are extracted from 11 individual sensors during isolated touch events and visualized using t-SNE~\cite{t-SNE}.
As shown in Fig.~\ref{fig:mtnet_manifold}a, the intrinsic tactile manifold forms clear and well-separated clusters that correspond to distinct spatial contact events. 
This separation arises from the sparse one-hot structure of the sensor representation, in which each contact activates only the corresponding sensor while the others remain near zero, resulting in consistently high silhouette scores (Fig.~\ref{fig:mtnet_manifold}b).
By contrast, shared image backgrounds and subtle pixel variations blur the visual distinctions among contact events, preventing clear geometric separation.

This discrepancy in manifold organization suggests the difficulty of mapping visual inputs to a structured tactile space.
In the untrained network, the silhouette score of the visual manifold is about 0.31 lower than that of the tactile manifold, indicating a diffuse latent space. Interestingly, optimization with reconstruction loss alone, without alignment constraints, further degrades manifold coherence and widens the silhouette score gap to about 0.66.
This deterioration may arise because direct regression encourages the network to converge toward average responses that minimize the overall numerical error.
Although the losses decrease as expected, distinct touch events become indistinguishable in the visual latent space, leaving the model with limited predictive capacity (Fig.~\ref{fig:supply_MTNetResult} and \ref{fig:supply_loss_curve}).
Only the full MTNet overcomes this collapse. 
Its visual manifold is reorganized into clear clusters that reflect the geometry of the tactile manifold. 
It achieves the highest Centered Kernel Alignment (CKA) score of about 0.74, outperforming the untrained and unaligned baselines by 0.22 and 0.55, respectively.

Pairwise cosine similarity matrices further corroborate the manifold observations (Fig.~\ref{fig:supply_MTNeSimMatrix}). 
The natural clustering within the tactile space is reflected by distinct diagonal patterns. 
In contrast, because the visual representations of the untrained and ablated models remain scattered, their similarity matrices appear uniformly saturated, resulting in little cross-modal correspondence. 
Only the fully constrained MTNet exhibits a structured visual matrix and establishes a meaningful diagonal correspondence across modalities. 
These findings highlight the necessity of the alignment mechanism. 
Given the substantial inherent discrepancy between vision and touch, a purely numerical mapping from pixels to forces is insufficient. 
By explicitly enforcing structural consistency, the alignment constraints regularize and simplify the visual-to-tactile mapping, encouraging the model to focus on contact-relevant physical structures, establishing a reliable foundation for visual-tactile inference.

\begin{figure}[tbp]
    \centering
    \includegraphics[width=0.99\textwidth]{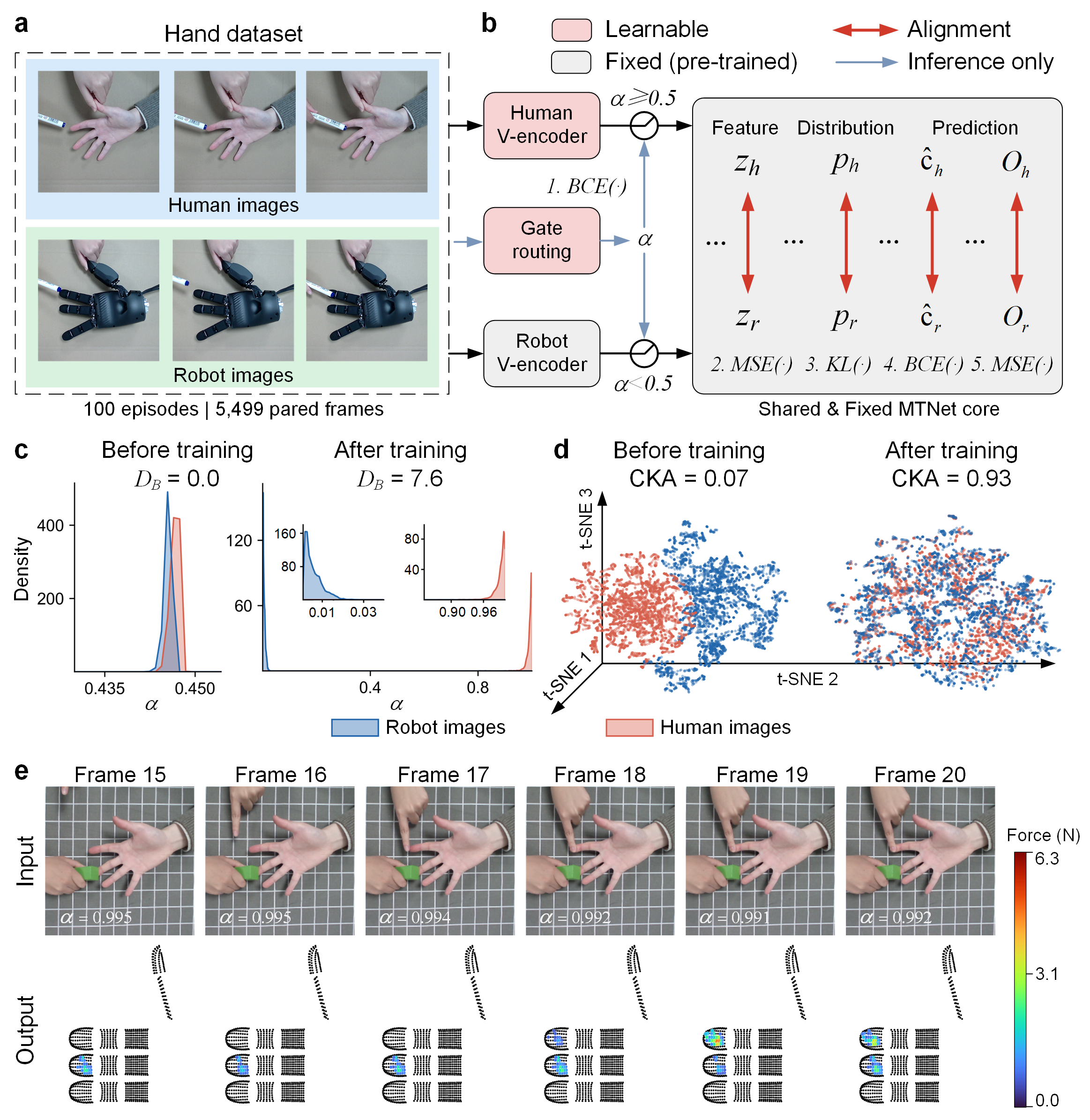}
    \caption{\textbf{AMTNet for human tactile prediction.}
    \textbf{a, }Hand dataset for training and evaluating AMTNet, constructed by manually matching human hand poses to robotic hand image frames from VT dataset.
    \textbf{b, }AMTNet architecture. AMTNet extends the pre-trained MTNet with a human visual encoder and a gating network, while keeping the MTNet core frozen. The gating network selects the domain-specific visual pathway. By aligning human visual space to the pre-trained robotic visual space, which is already aligned with the robotic tactile space, AMTNet transfers learned robotic visuo-tactile prediction to human hand images (see Methods for details).
    \textbf{c, }Distributions of the gating output $\alpha$ before and after training, where $\alpha \rightarrow 1$ denotes the human images and $\alpha \rightarrow 0$ denotes the robotic images.
    \textbf{d, }t-SNE visualizations of human hand and robotic hand visual features before and after training.
    \textbf{e, }Representative AMTNet predictions from human hand images.
    }
    \label{fig:amtnet}
\end{figure}

\subsection{Aligned generalization to human hands}
\label{sec:amtnet}
We further extend the alignment principle of MTNet to bridge the morphological gap between human and robotic hands, enabling tactile prediction from human visual input. 
A key challenge is the absence of ground-truth tactile data for human hands, which precludes supervised training of a human tactile predictor.
This setting resembles observation-based perceptual generalization in humans.
To address this, we explore a cross-domain knowledge transfer approach based solely on visual space alignment and construct a paired dataset of human and robotic hands (Fig.~\ref{fig:amtnet}a). 
Using the robotic RGB images in VT dataset as reference, we collect corresponding human hand images by reproducing the hand poses and interaction content in those episodes, yielding a paired dataset of 100 interaction sequences and 5,499 image frame pairs. 
On this basis, our cross-domain generalization strategy aligns the human visual space with the established robotic visual space, and thereby indirectly with the robotic tactile space.

We introduce AMTNet, an extension of the pre-trained MTNet for cross-domain alignment (Fig.~\ref{fig:amtnet}b).
AMTNet incorporates an additional human-specific visual encoder and a gating network, which routes input images to either the human or robotic visual encoder according to its domain. 
During training, the visual encoder, tactile encoder, and decoder inherited from MTNet are frozen to preserve the established alignment between the robotic visual and tactile spaces and avoid catastrophic forgetting. 
Training is therefore restricted to the human visual encoder and gating network. 
The objective enforces cross-domain alignment across visual feature representations, latent prior distributions, predicted contact states, and tactile reconstructions.
In this way, the human visual domain is embedded into the robotic visual manifold already aligned with the tactile space, enabling the frozen MTNet decoder to transfer robotic visual-tactile prediction to human hand images.

Experimental results demonstrate the effectiveness of AMTNet in cross-domain alignment and prediction. 
The gating network accurately discriminates between human and robotic inputs. 
Its output distributions initially overlap but become clearly separated after training, concentrating near opposite ends associated with the two domains, with the bhattacharyya distance ($D_b$)~\cite{Bhattacharyya} increasing from 0 to 7.6 (Fig.~\ref{fig:amtnet}c). 
Visualizations of the representation space show that initially distinct human and robotic visual representations converge onto a shared latent manifold after training, with the CKA similarity between the two feature domains increasing from 0.07 to 0.93 (Fig.~\ref{fig:amtnet}d).
As shown in Fig.~\ref{fig:amtnet}e, by leveraging this core alignment principle, AMTNet directly predicts detailed robotic tactile signals from human hand images, including contact locations and force magnitudes.
The predictions are consistent with those obtained from the corresponding robotic hand images (Fig.~\ref{fig:supply_AMTNetResult}), demonstrating cross-domain tactile prediction without requiring human tactile supervision or direct visual-to-tactile regression training in the human domain.

\subsection{Robotic implementation of mirror touch}
\label{sec:mirrot_touch}
\begin{figure}[tbp]
    \centering
    \includegraphics[width=0.99\textwidth]{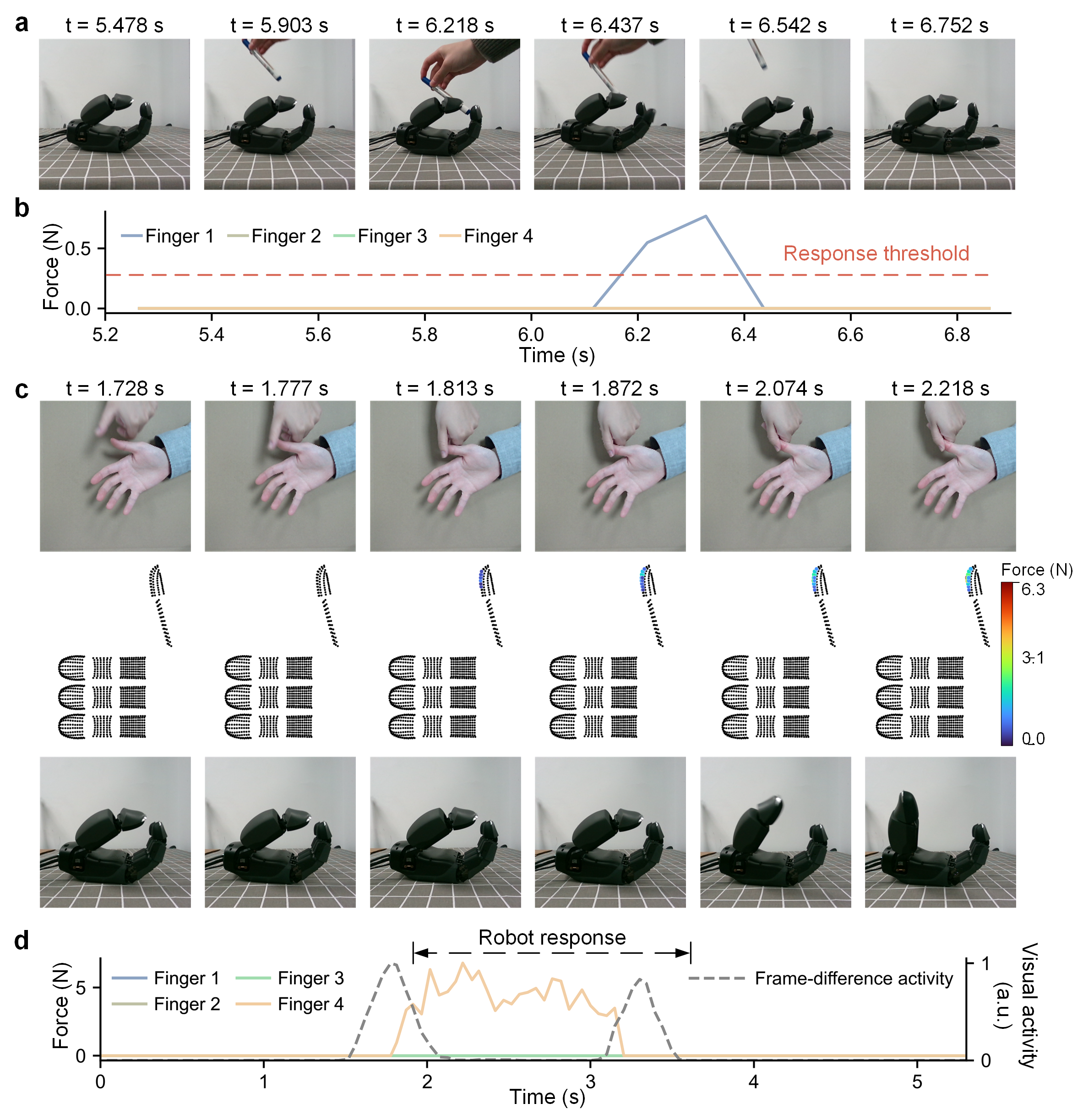}
    \caption{\textbf{Robotic mirror touch with trained AMTNet.}
    \textbf{a, }Robotic tactile reflex setup, in which a predefined controller triggered a flick response when the maximum force magnitude for the corresponding finger exceeded 0.2 N and reset after 0.5 s without tactile input.
    \textbf{b, }Changes in maximum force magnitude across four robotic fingers in the contact example shown in \textbf{a}.
    \textbf{c, }Mirror touch demonstration in which AMTNet predicts tactile signals from 30 Hz images of a touched human hand, eliciting corresponding finger responses in a physically untouched robotic hand.
    \textbf{d, }Predicted four-finger tactile signals in \textbf{c, }overlaid with the frame-difference motion curve of the hand images to illustrate their temporal correspondence with visual touch motion.
    See Fig.~\ref{fig:supply_fingerResponse} and Supplementary Video 3 for unseen object and extended demonstrations.
    }
    \label{fig:finger}
\end{figure}

Finally, we evaluate whether AMTNet predictions could drive behavior resembling mirror touch on a physical robotic hand. 
To this end, we conduct a real-robot experiment in which the system was programmed to generate a physical response whenever the maximum contact force on any robotic finger exceeded a preset threshold of 0.2 N.
Once the threshold is crossed, the corresponding finger executes a flick response. 
As a baseline test (Fig.~\ref{fig:finger}a-b), direct physical interaction with the robotic hand reliably elicited this predefined response. 
The recorded force traces confirm that the peak force on the contacted finger rapidly exceeds the threshold, demonstrating the hardware system can generate natural touch-triggered responses.

We then examine whether the same response can be elicited by visual observation alone. 
In this experiment (Fig.~\ref{fig:supply_setup}a), a human hand is touched under the camera, while the robotic hand receives no physical contact. 
Nevertheless, AMTNet predicts tactile signals from the visual input, and these predictions are sufficient to trigger flick responses in the corresponding robotic fingers (Fig.~\ref{fig:finger}c). 
The predicted tactile trajectories for the four fingers are shown in Fig.~\ref{fig:finger}d. 
For comparison, we also compute frame-difference motion traces from the human hand images. 
The close temporal alignment between the predicted tactile signals and the visual motion traces indicates that the robotic hand generates tactile responses in synchrony with the observed touch events. 
Supplementary Video 3 and Fig.~\ref{fig:supply_fingerResponse}c-d provide an intuitive overview of visually evoked robotic responses elicited by touches on different regions of the human hand, including those objects unseen in both VT and Hand datasets. 
This robustness can be attributed to the alignment process, which guides the model to focus on the core contact event while reducing interference from irrelevant visual cues.

Collectively, these components establish a complete process from the alignment principle to embodied visuo-tactile resonance.
MTNet first learns a structurally aligned correspondence between vision and touch, and AMTNet then maps visual features of another hand into this shared representation space. 
This mapping enables visual observations of human touch to drive self-referenced tactile prediction and physical robotic responses, demonstrating that detailed tactile signals can be synthesized without tactile recordings from the observed hand.

\section{Discussion}\label{sec:discussion}
In this study, we show that a computational framework inspired by the cortical alignment principle can support visuo-tactile resonance in a robotic system. 
This framework comprises MTNet and its adaptive extension AMTNet, which impose multi-level structural alignment across feature semantics, latent distributions, and sample-wise distances, allowing the system to predict detailed tactile signals from visual input alone. 
Aligning visual and tactile representations supports not only cross-modal prediction but also cross-domain generalization from the robot hand to the human hand, such that observed human touch can elicit tactile responses in the robot fingers. 
This study connects neuroscientific accounts of mirror touch with their implementation in embodied systems and shows that a robot can perceive what it sees by aligning its internal sensory manifold with the observed experiences of others.

Our work complements prior studies on visuo-tactile integration but addresses a more heterogeneous form of cross-modal learning. 
Mainstream approaches often use vision-based tactile grippers~\cite{OmniVTA,TacVisionMani_2025arxiv,TacMani_2025CoRL,tac_pred_MAE_2025CoRL,tac_pred_MAE_2025ICLR,vt_cl_2025ICRA}, where mappings are typically learned between two image-like domains. 
Although valuable for tactile perception, these systems weakly reflect the human-like modality gap between vision and touch. 
By contrast, our tactile signals are measured at millimetre resolution over a densely distributed sensing surface on a dexterous robotic hand, with contact activations sparse in both space and time. 
While better capturing the structure of hand-based touch, this setting makes visual-to-tactile mapping more heterogeneous and underdetermined, motivating a departure from current tactile prediction methods that mainly rely on reconstruction objectives for end-to-end training~\cite{OmniVTA,tac_pred_MAE_2025ICLR,tac_pred_MAE_2026SciRob,tac_pred_MAE_2025CoRL}. 
Our approach instead emphasizes explicit multi-level alignment constraints that enforce structural consistency between visual and tactile modalities, and then leverages this principle to transfer tactile prediction capability to the human hand in the absence of human tactile references. 
In this sense, the framework more closely resembles human learning~\cite{human_learning_2025NC}, in which novel scenarios are merged into existing concepts to support knowledge generalization.

Several limitations should also be noted.
First, because contact events inevitably induce occlusion, tactile states cannot be fully recovered from monocular RGB input alone, particularly fine-grained 3D force values at the taxel level. 
Current systems are therefore more reliable at coarse contact locations and force magnitudes, although these estimates remain valuable for meaningful behaviour. 
Second, the cross-domain transfer in AMTNet depends on the quality of paired human-robotic hand images, and its robustness is limited when contact poses and motion speeds differ substantially across domains. 
Introducing a small number of human tactile ground-truth samples as anchors may further improve this transfer. Third, the behavioural expression examined here is intentionally simple: a threshold-triggered flick response. 
This design helps isolate the contribution of tactile prediction, but does not yet link mirror touch experience to richer motor or cognitive responses.

These limitations point to several directions for future work.
One direct extension is to apply the alignment framework to a wider range of hand morphologies, contact conditions and sensory modalities, so that the learned visuo-tactile resonance can support more complex forms of cross-body generalization. 
Another is to combine alignment-based tactile inference with higher-level motor control and cognitive models, allowing predicted tactile signals to inform action planning and human-robot interaction. 
More broadly, our results suggest that explicit alignment across perceptual spaces can enable embodied systems to respond to human tactile experiences.
Extending this principle beyond the present hand-based tactile setting may contribute to more general forms of embodied alignment between humans and robots.



\bmhead{Acknowledgements}
This work was supported by the Tsinghua University–Xiamen Dalle New Energy Automobile Co., Ltd. Joint Institute for Embodied and Collaborative Robotics, Dalle, D-Robotics, and the National Natural Science Foundation of China grants (T2122015).

\bibliography{sn-bibliography}
\clearpage
\section*{Appendix}
\renewcommand{\figurename}{Fig.}
\setcounter{figure}{0}
\renewcommand{\thefigure}{S\arabic{figure}}
\begin{figure}[hbp]
    \centering
    \includegraphics[width=0.8\textwidth]{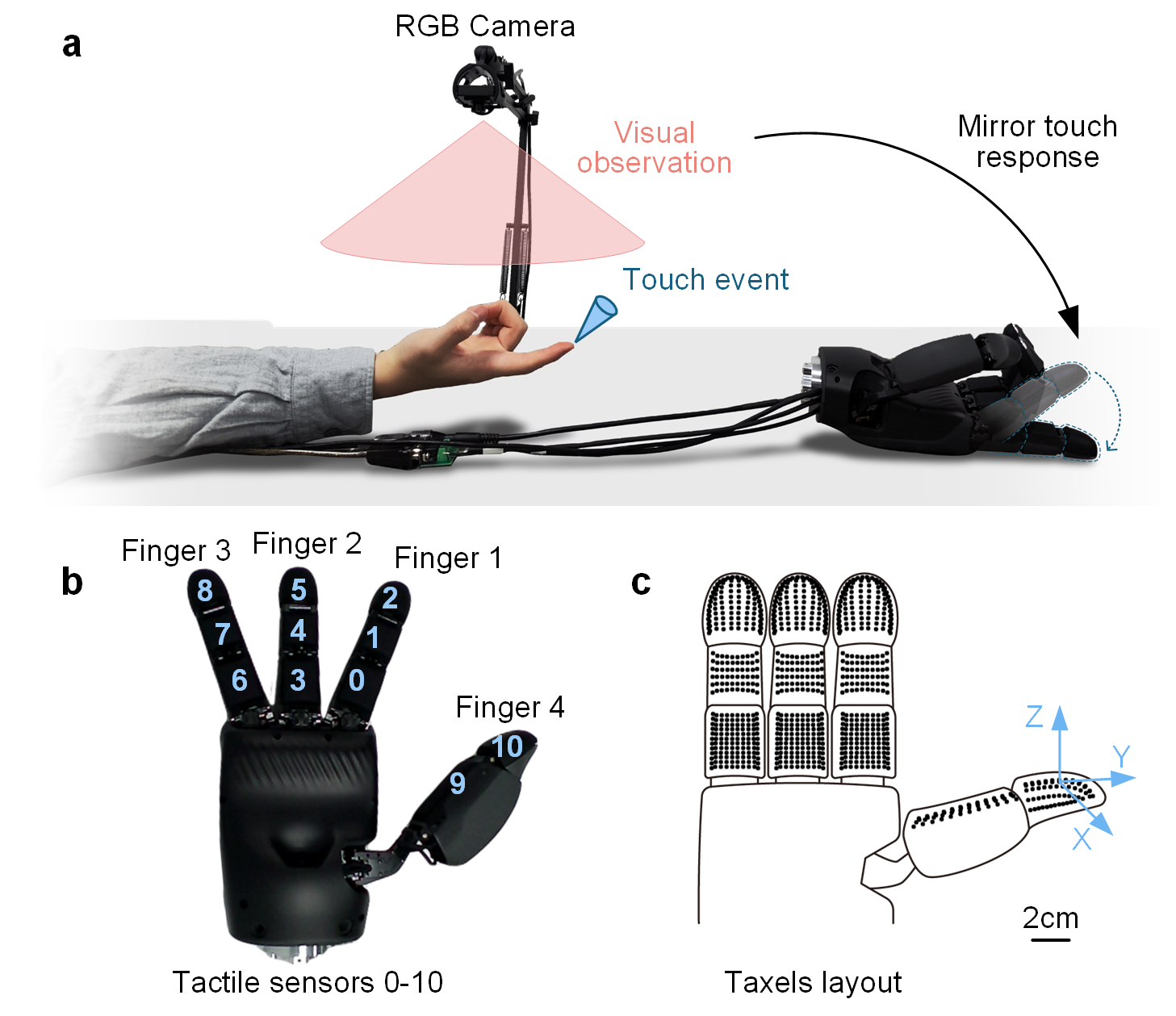}
    \caption{\textbf{Hardware setup.}
    \textbf{a, }The robotic mirror touch setup uses an RGB camera to capture touch events on a human hand and predict the corresponding tactile responses of the robotic hand.
    \textbf{b-c, }The robotic hand contains 11 electromagnetic tactile sensors with 1,140 independent taxels in total, providing 1-mm spatial resolution and three-axis force sensing.
    }
    \label{fig:supply_setup}
\end{figure}

\clearpage
\renewcommand{\figurename}{Fig.}
\setcounter{figure}{1}
\renewcommand{\thefigure}{S\arabic{figure}}
\begin{figure}[tbp]
    \centering
    \includegraphics[width=0.99\textwidth]{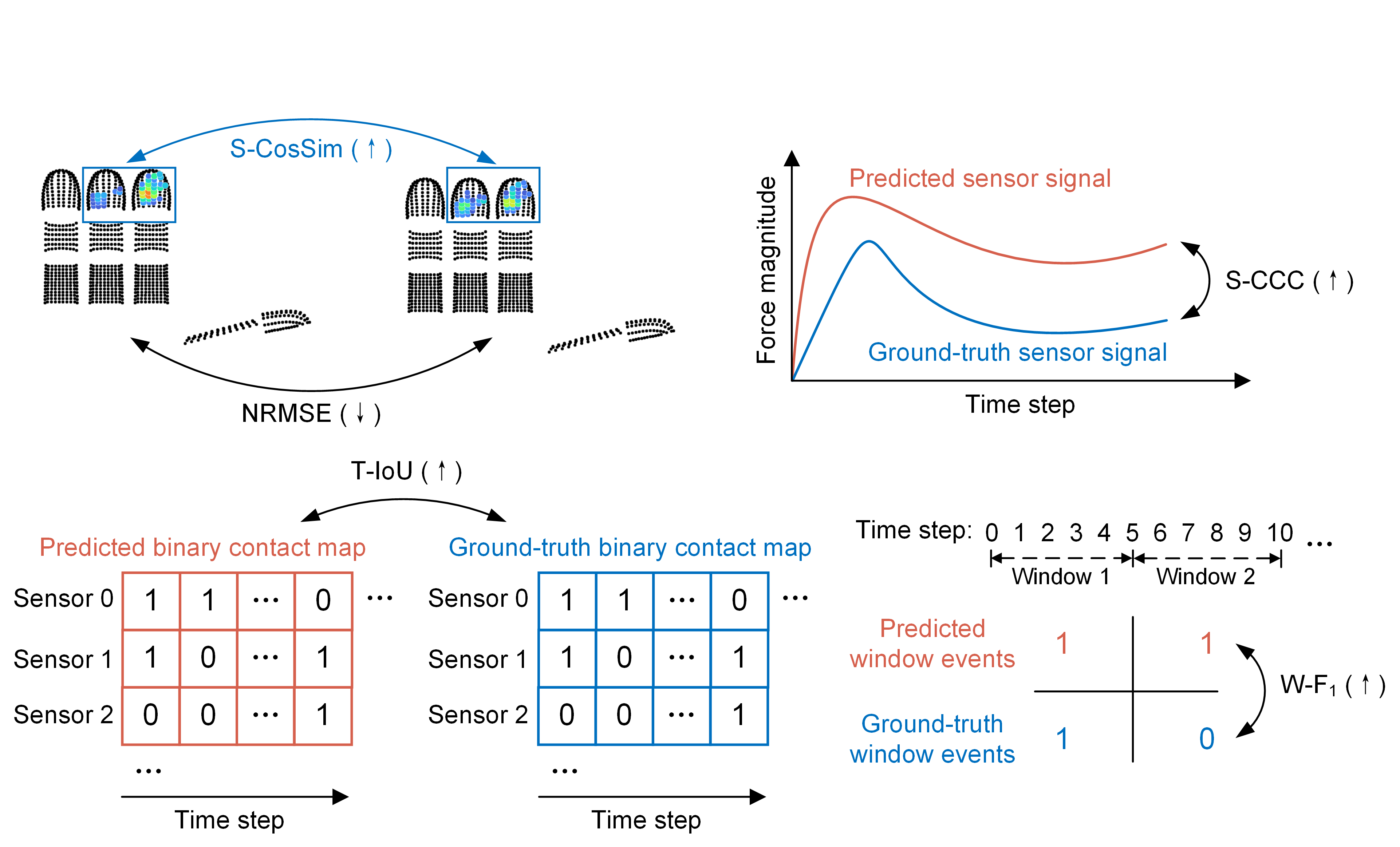}
    \caption{\textbf{Schematic of MTNet evaluation metrics.}
    \textbf{NRMSE} quantifies the force magnitude error over the full taxel fields.
    \textbf{S-CosSim} measures the spatial similarity of tactile force distributions within contact regions.
    \textbf{S-CCC} evaluates the temporally concordance at the sensor-level signals.
    \textbf{T-IoU} assesses the spatial and temporal overlap between thresholded contact maps.
    \textbf{W-F1} measures contact event detection accuracy within temporal windows, emphasizing event-level timing consistency.
    See Sec.~\ref{sec:metrics} for details.
    } 
    \label{fig:supply_metrics}
\end{figure}

\renewcommand{\thetable}{S\arabic{table}}
\begin{table}
\caption{
\textbf{Ablation on alignment constraints in MTNet.}
Unaligned/Aligned denote MTNet trained without/with the three alignment constraints in Fig.~\ref{fig:mtnet}d-h; w/o indicates removal of the corresponding constraint. Results are mean $\pm$ SD over five runs.
}
\label{tab:alignment_ablation}
\setlength{\extrarowheight}{10pt}
\begin{tabular}{
>{\raggedright\arraybackslash}m{3.5cm}
>{\centering\arraybackslash}m{1.4cm}
>{\centering\arraybackslash}m{1.4cm}
>{\centering\arraybackslash}m{1.4cm}
>{\centering\arraybackslash}m{1.4cm}
>{\centering\arraybackslash}m{1.4cm}
}
\toprule
Model variant & \makecell{NRMSE \\$(\downarrow)$} & \makecell{S-CosSim \\$(\uparrow)$}& \makecell{T-IoU \\$(\uparrow)$}& \makecell{W-F1 \\$(\uparrow)$}& \makecell{S-CCC \\$(\uparrow)$}\\
\midrule
Unaligned MTNet & \makecell{0.0576 $\pm$\\ 0.0023} & \makecell{0.1682 $\pm$\\ 0.0059} & \makecell{0.0090 $\pm$\\ 0.0067} & \makecell{0.0816 $\pm$\\ 0.0003} & \makecell{0.1397 $\pm$\\ 0.0001} \\
\makecell[l]{MTNet w/o \\distribution alignment} & \makecell{0.0640 $\pm$\\ $<$ 0.0001} & \makecell{0.2217 $\pm$\\ 0.0020} & \makecell{0.0404 $\pm$\\ 0.0006} & \makecell{0.0818 $\pm$\\ $<$ 0.0001} & \makecell{0.1397 $\pm$\\ $<$ 0.0001} \\
\makecell[l]{MTNet w/o \\representational alignment} & \makecell{0.0101 $\pm$\\ $<$ 0.0001} & \makecell{0.5921 $\pm$\\ 0.0010} & \makecell{0.8669 $\pm$\\ 0.0007} & \makecell{0.7364 $\pm$\\ 0.0021} & \makecell{0.9524 $\pm$\\ 0.0009} \\
\makecell[l]{MTNet w/o \\rational alignment} & \makecell{\textbf{0.0098} $\pm$\\\textbf{$<$ 0.0001}} & \makecell{0.6078 $\pm$\\ 0.0015} & \makecell{0.8710 $\pm$\\ 0.0005} & \makecell{0.7573 $\pm$\\ 0.0017} & \makecell{0.9553 $\pm$\\ 0.0004} \\
Aligned MTNet & \makecell{0.0102 $\pm$\\ $<$ 0.0001} & 
\makecell{\textbf{0.6097} $\pm$ \\\textbf{0.0010}} & \makecell{\textbf{0.8814}  $\pm$\\\textbf{0.0005}} & \makecell{\textbf{0.7920} $\pm$\\\textbf{0.0019}} & \makecell{\textbf{0.9664} $\pm$ \\\textbf{0.0007}} \\
\bottomrule
\end{tabular}
\end{table}

\renewcommand{\figurename}{Fig.}
\setcounter{figure}{2}
\renewcommand{\thefigure}{S\arabic{figure}}
\begin{figure}[tbp]
    \centering
    \includegraphics[width=0.99\textwidth]{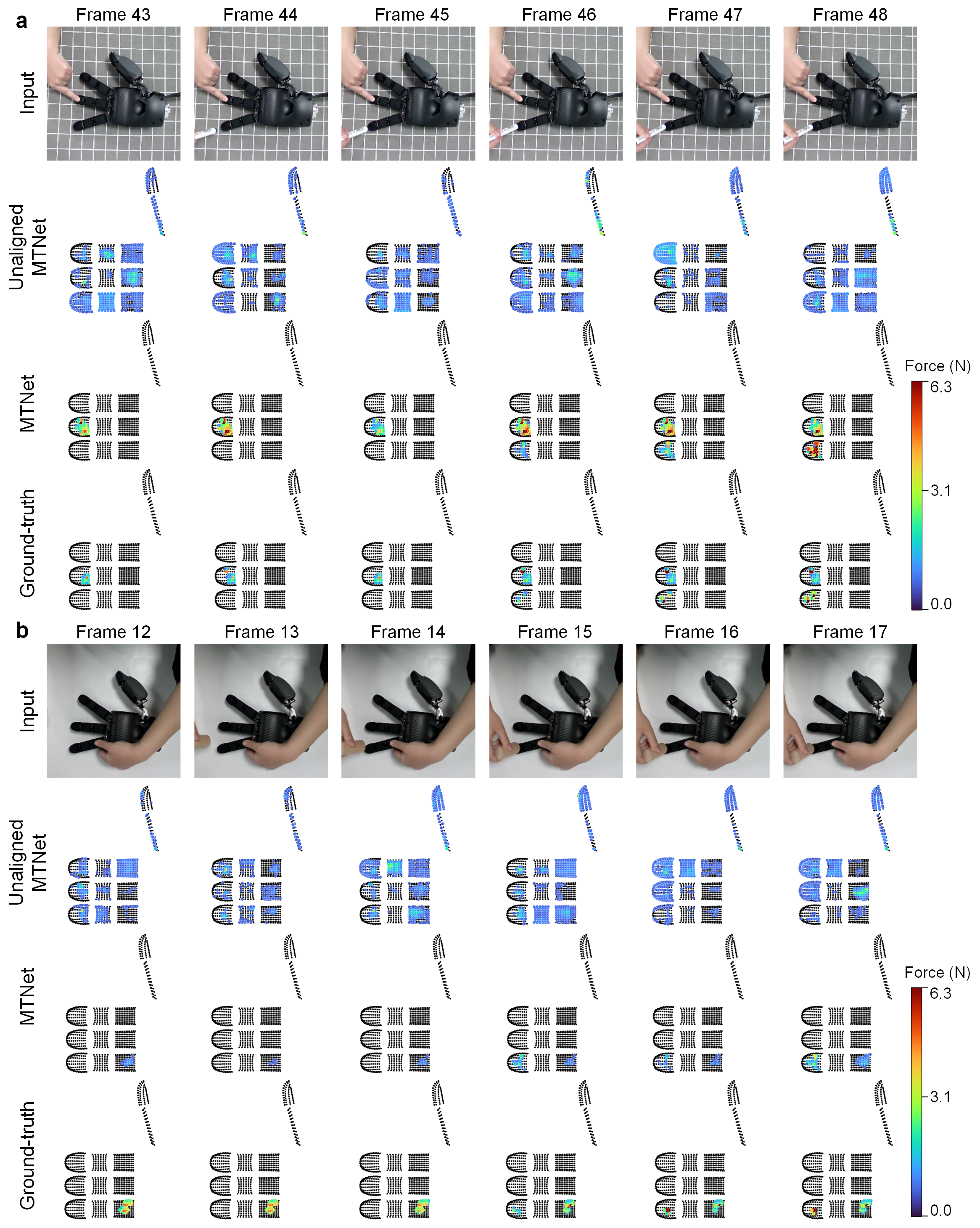}
    \caption{\textbf{Tactile prediction results of MTNet with and without alignment constraints.}
    \textbf{a, }Expanded from Fig.~\ref{fig:mtnet_result}b, showing MTNet without alignment constraints. 
    \textbf{b, }Same as \textbf{a} for another sample segment.
    } 
    \label{fig:supply_MTNetResult}
\end{figure}

\renewcommand{\figurename}{Fig.}
\setcounter{figure}{3}
\renewcommand{\thefigure}{S\arabic{figure}}
\begin{figure}[tbp]
    \centering
    \includegraphics[width=0.99\textwidth]{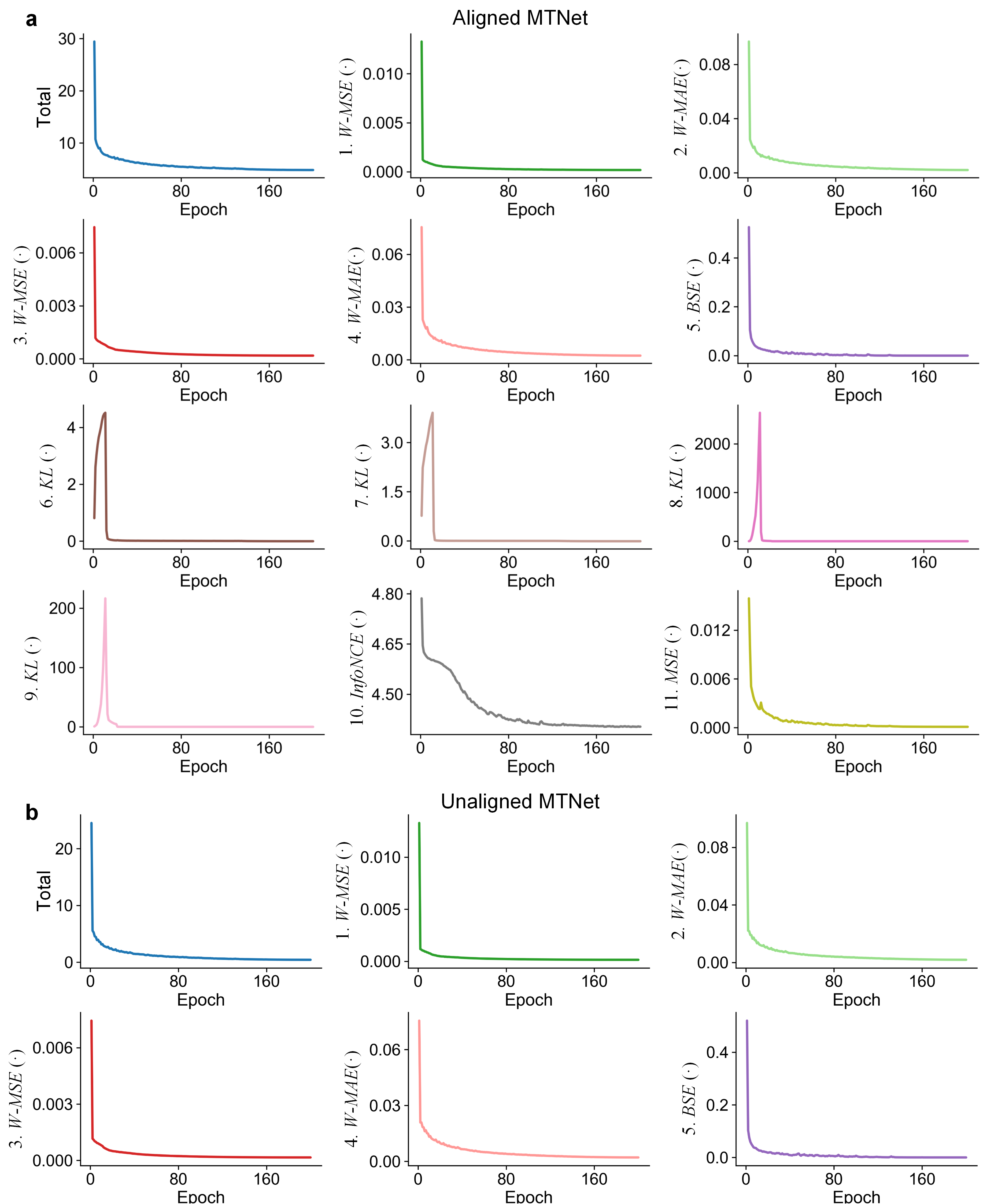}
    \caption{\textbf{Training loss curves of MTNet.}
    \textbf{a, }Eleven losses of the fully trained MTNet, ordered as in Fig.~\ref{fig:mtnet}d-h, with the first five corresponding to reconstruction losses and the remaining six to alignment constraints. 
    \textbf{b, }Losses of the unaligned MTNet trained with only the five reconstruction losses.
    } 
    \label{fig:supply_loss_curve}
\end{figure}

\renewcommand{\figurename}{Fig.}
\setcounter{figure}{4}
\renewcommand{\thefigure}{S\arabic{figure}}
\begin{figure}[tbp]
    \centering
    \includegraphics[width=0.80\textwidth]{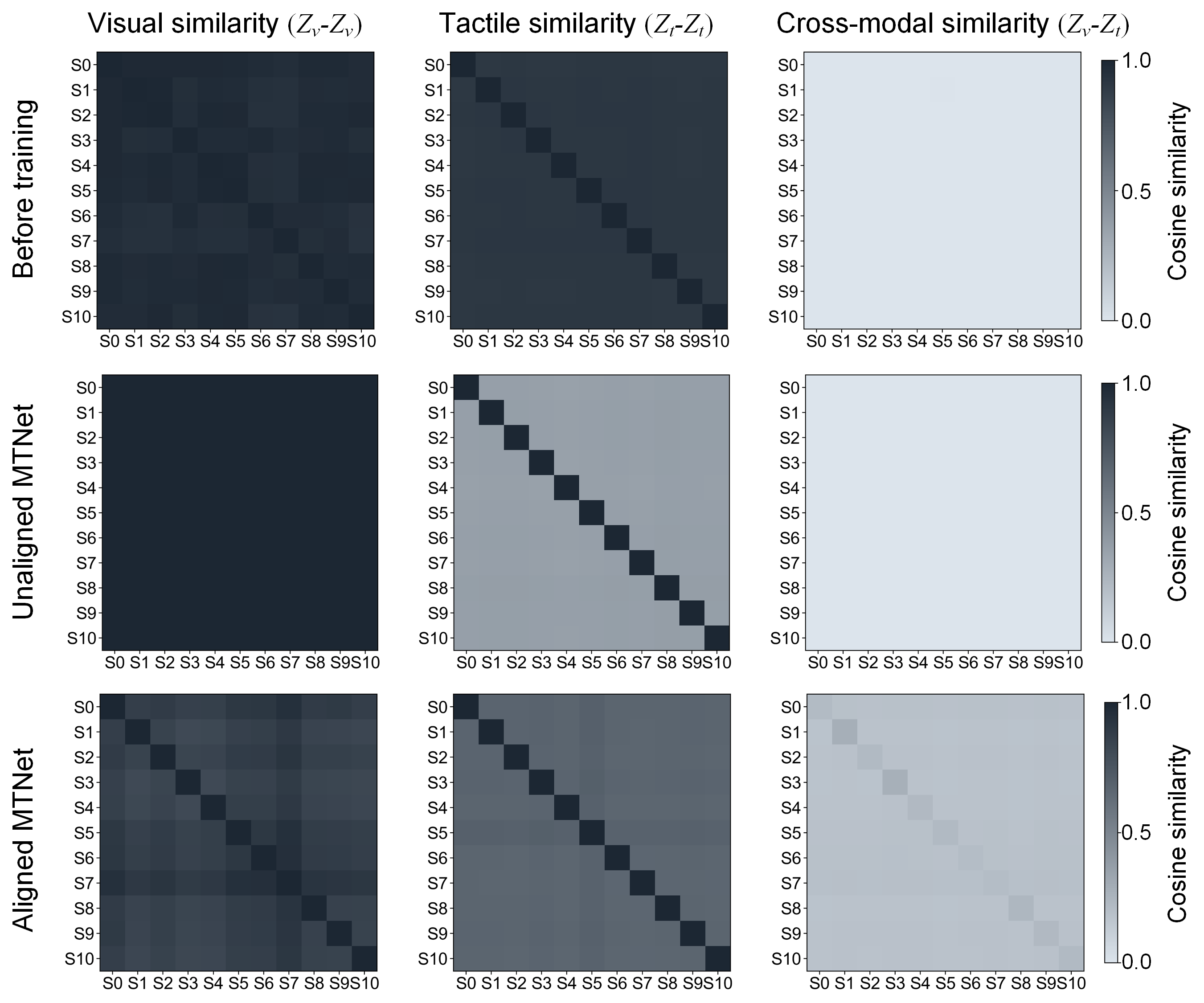}
    \caption{\textbf{Latent representation similarity of MTNet.}
    Based on the single sensor touch setting for 11 sensors (S0–S10) in Fig.~\ref{fig:mtnet_manifold}a, cosine similarity matrices are computed for the $Z_v$ and $Z_t$ representations of untrained MTNet, MTNet without alignment constraints, and fully trained MTNet. From left to right: $Z_v \text{-}Z_v$, $Z_t \text{-}Z_t$, and $Z_v \text{-}Z_t$ similarity matrices.
    } 
    \label{fig:supply_MTNeSimMatrix}
\end{figure}

\renewcommand{\figurename}{Fig.}
\setcounter{figure}{5}
\renewcommand{\thefigure}{S\arabic{figure}}
\begin{figure}[tbp]
    \centering
    \includegraphics[width=0.99\textwidth]{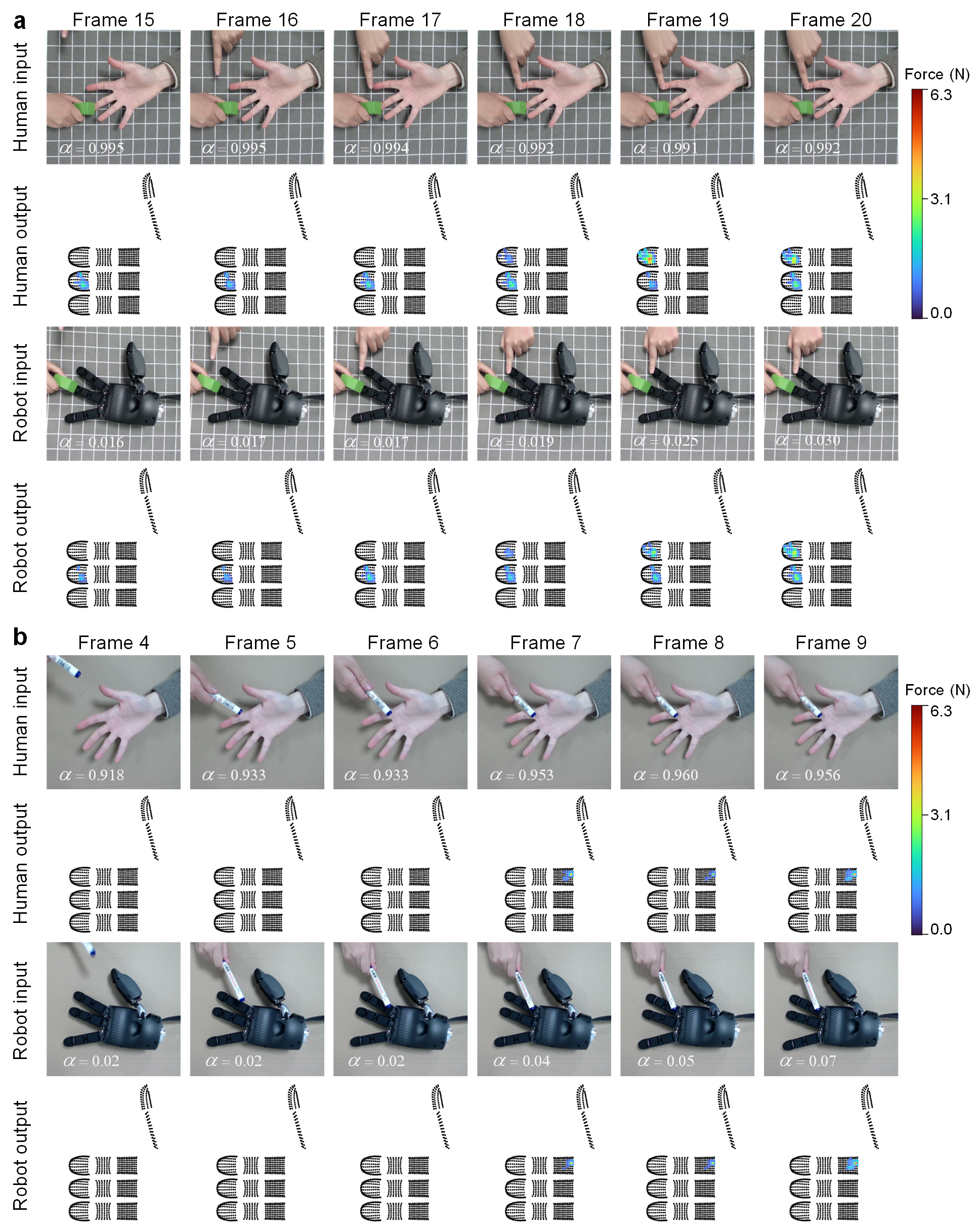}
    \caption{\textbf{Representative prediction results of AMTNet for human hand and robotic hand images.}
    \textbf{a, }Expanded from Fig.~\ref{fig:amtnet}e, including the corresponding prediction results for robotic images. 
    \textbf{b, }Same as \textbf{a} for another sample segment.
    } 
    \label{fig:supply_AMTNetResult}
\end{figure}

\renewcommand{\figurename}{Fig.}
\setcounter{figure}{6}
\renewcommand{\thefigure}{S\arabic{figure}}
\begin{figure}[tbp]
    \centering
    \includegraphics[width=0.99\textwidth]{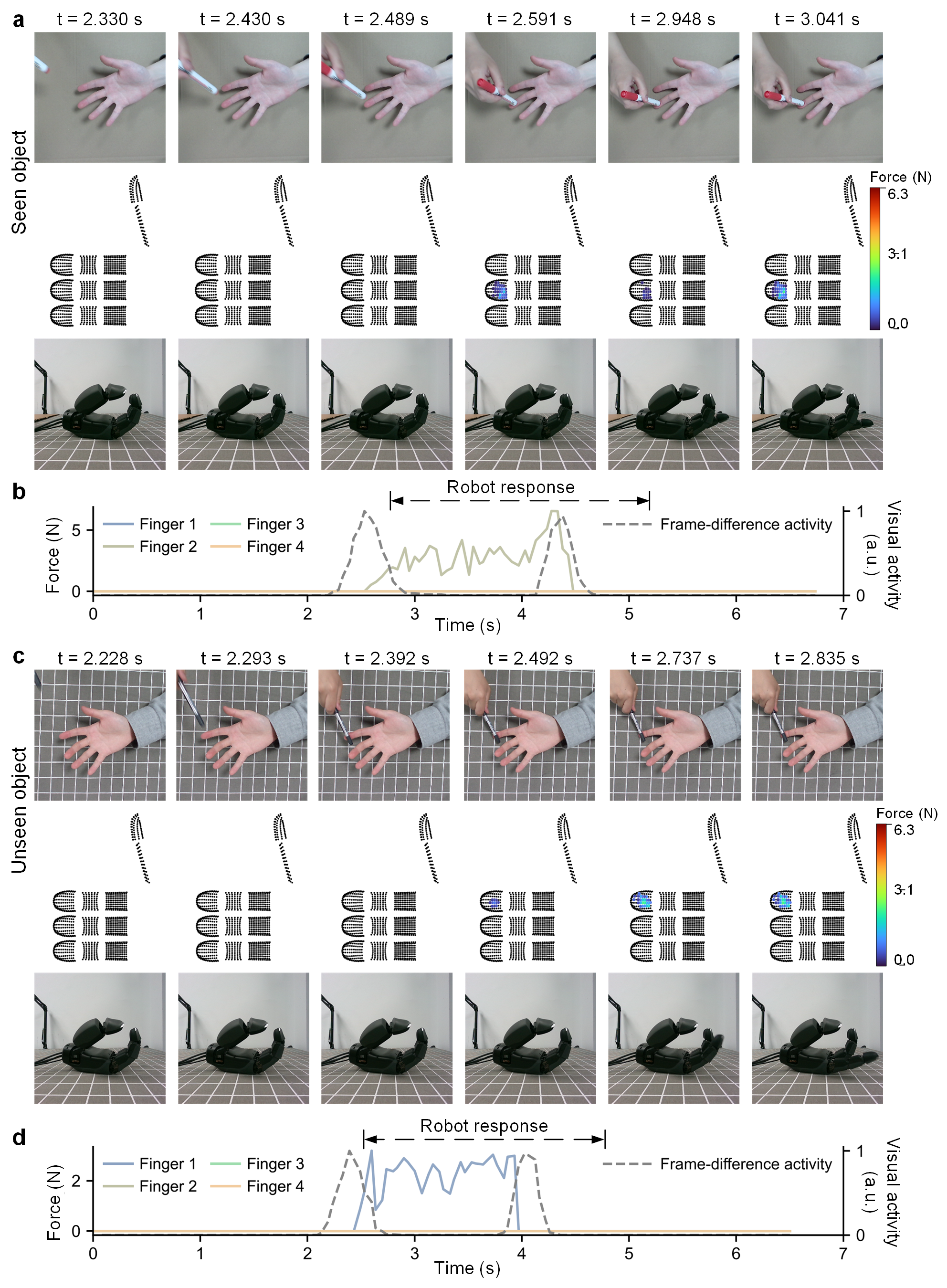}
    \caption{\textbf{a, }Visually evoked robotic tactile responses induced by the object seen in VT and Hand datasets.
    \textbf{b, }Predicted four-finger tactile signals and frame-difference motion curve corresponding to \textbf{a}. 
    \textbf{c-d, }Same as \textbf{a-b}, but for unseen objects.
    Full examples are provided in Supplementary Video 3.
    } 
    \label{fig:supply_fingerResponse}
\end{figure}



\end{document}